\documentclass{article}
\usepackage{graphicx}
\usepackage[top=1in, bottom=1in, left=1in, right=1in]{geometry}
\usepackage[fleqn]{amsmath}
\usepackage{caption}
\usepackage{subfig}
\usepackage{amsthm}
\date{}

\begin{document}
	\title{FHEDN: A based on context modeling Feature Hierarchy Encoder-Decoder Network for face detection}
	\author{\small Zexun Zhou\textsuperscript{1}, Zhongshi He\textsuperscript{1}, Ziyu Chen\textsuperscript{1}, Yuanyuan Jia\textsuperscript{2}, Haiyan Wang\textsuperscript{1}, Jinglong Du\textsuperscript{1}, Dingding Chen\textsuperscript{1}\\
		\small College of Computer Science,Chongqing University,Chongqing,China\textsuperscript{1}\\
		\small College of Medical Informatics,Chongqing Medical University,Chongqing,China\textsuperscript{2}\\
		\small Sichuan Fine Arts Institute,Chongqing,China\textsuperscript{3}\\
		\small \{zexunzhou,zshe,chenziyu,yyjia,jldu,dingding\}@cqu.edu.cn;\{why\}@scfai.edu.cn}

	\maketitle
	
	\begin{abstract}
		Because of affected by weather conditions, camera pose and range, etc. Objects are usually small, blur, occluded and diverse pose in the images gathered from outdoor surveillance cameras or access control system. It is challenging and important to detect faces precisely for face recognition system in the field of public security. In this paper, we design a based on context modeling structure named Feature Hierarchy Encoder-Decoder Network for face detection(FHEDN), which can detect small, blur and occluded face with hierarchy by hierarchy from the end to the beginning likes encoder-decoder in a single network. The proposed network is consist of multiple context modeling and prediction modules, which are in order to detect small, blur, occluded and diverse pose faces. In addition, we analyse the influence of distribution of training set, scale of default box and receipt field size to detection performance in implement stage. Demonstrated by experiments, Our network achieves promising performance on WIDER FACE and FDDB benchmarks. 
	\end{abstract}

\section{Introduction}
\ \ \ Face detection is most studied in computer vision recent years. It is a core module of face recognition system which has successfully applied in many areas such as public security surveillance, smart pay, etc. Many state-of-the-art algorithm for face detection have been present over the past two decades.

Previous methods utilize hand-craft feature and specific classifier to detect face from any natural images. Viola and Jones\cite{Viola2003Rapid} put forward the cascade face detector based on Haar-like feature and AdaBoost classifier, whose excellent real-time detection performance made it milestone in face detection. Followed by Viola and Jones, many improved work based on extent Haar-like feature and cascade boosting based methods have been proposed. Dong et al. designed a joint cascaded framework named JDA for face detection and alignment\cite{Chen2014Joint}, which adopt shape indexed feature and cascade boosting based classifier to solve the two tasks jointly. Shengcai et al. proposed a fast and accurate unconstrained face detector which took advantage of feature named Normalized Pixel Difference(NPD) and deep quadratic tree structure\cite{Liao2016A}. Besides, due to the state-of-the-art result by Deformable Parts Models(DPMs) in object detection\cite{Felzenszwalb2010Object}, DPMs also successfully applied for face detection and achieved competitive performance compared with other method\cite{Mathias2014Face,Ramanan2012Face}.
 
With the state-of-the art performance gained by DCNNs in computer vision, more and more based on DCNNs models have been proposed for generic object detection task. They can be categorized into two categories. One is scale-invariant methods, such as the seminal work of RCNN\cite{Girshick2015Region}, Fast RCNN\cite{Girshick2015Fast} and Faster-RCNN\cite{Ren2017Faster}, etc. The other is scale-variant methods which includes YOLO\cite{Joseph2015You}, SSD\cite{Liu2015SSD}, etc. Recent face detection methods typically follow the paradigm of the two categories. These methods use DCNNs as the backbone structure to learn highly discriminative representation. Among these methods, \cite{Li2015A,Qin2016Joint,Zhang2016Joint} combined traditional cascade style with region proposal into DCNNs and achieve a good trade-off between accuracy and speed, in contrast, \cite{Hu2016Finding,Yang2017Face,Zhang2017S,Najibi2017SSH} follow methodology of scale-variant such that they can detect face in multiple feature hierarchies without constructing image pyramids.  

In this paper, we focus on leveraging Deep Convolutional Neural Networks(DCNNs) to detect small and blur face in unconstrained scene. As we known, small object detection is an open and challenging problem. But the images gathered from outdoor survillance cameras, which are usually affected by weather conditions, camera pose and range, etc. Consequently, objects in the images are small, blur, occluded and diverse pose. How to detect face above described well in surveillance camera has been a key problem for subsequent face recognition. Therfore, we are interested in constructing a framework based on DCNNs for solved above problem. 

This paper mainly makes following contributions:\\
(1) Inspired by the idea of scale-variant methods such as Single Shot Detector(SSD)\cite{Liu2015SSD}, we firstly fine-tuned SSD for face detection task with Annotated Facial Landmarks in the Wild(AFLW)\cite{K2012Annotated} as training dataset. We then analyzed the weak capability about detecting small face of trained model from distribution characteristics of training dataset.\\
(2) We designed a based on context modeling Feature Hierarchy Encoder-Decoder Network(FHEDN) for face detection to detect small and blur face in unconstrained scene, which learns and fuses context information around the face adjacent to the previous feature hierarchy. Due to face in image is not independent individual and it is both at top of the neck and bottom of the hair. These important context information around face can help FHEDN to improve detection performance.\\
(3) For modeling context by FHEDN, we employed a stacked hourglass network structure\cite{Newell2016Stacked} to fuse context information together with hierarchy-wise. Therefore, the overall network adopts scale-variant designing paradigm which likes "encoder-decoder" style. Furthermore, we analyzed the influence of default box scale, receptive field size etc. on detection performance(discussed in section 4.2). 

\section{Related Work}
\ \ \ As the survey on face detection elaborated thoroughly\cite{Zafeiriou2015A}, there are numerous works in the field of face detection. In this section, we only focus on a series of work exploiting deep learning. 

Early in 1900s, there are existing works using neural networks for face detection task. Vaillant et al.\cite{Vaillant1993Original} proposed a two-stage CNNs to detect faces from images in coarse-to-fine manner. Rowley et al.\cite{Rowley1998Rotation} presented a approach based on CNN for upright frontal face detection and achieved improved performance. Garcia et al.\cite{Christophe2002A} developed a convolutional neural architecture for different pose and rotation angles to detect faces. In 2005 Osadchy et al.\cite{Osadchy2004Synergistic} designed a multi-task neural network to train jointly face detector and pose estimator. In recent five years, deep learning has been pay more and more attention because of its state-of-the-art performance in the area of computer vision, natural language processing and speech recognition, etc. In various deep neural networks architecture of deep learning, due to DCNNs have achieved breakthrough results in computer vision task such as image classification\cite{Krizhevsky2012ImageNet,Zeiler2014Visualizing,Simonyan2014Very,Szegedy2014Going,He2015Deep}, genetic object detection\cite{Girshick2015Region,Girshick2015Fast,Ren2017Faster,Joseph2015You,Liu2015SSD,Sermanet2013OverFeat,Uijlings2013Selective,He2015Spatial,Najibi2016G} and semantic segmentation\cite{Noh2015Learning,Jonathan2017Fully}, and thus it been mostly studied. Inspired by deep learning-based methods in generic object detection, face detection algorithm based on DCNNs can be categorized into tow categories: scale-invariant methods and scale-variant methods. 

Scale-invariant methods: The seminal work of Faster RCNN employs region of interest(ROI) pooling to extract scale-invariant features. In addition, through traditional cascaded classification scheme, some new face detection algorithm based on DCNNs have been proposed in recent years. Haoxiang et al.\cite{Li2015A} put forward a convolutional neural network for face detection, which consists of two stage cascaded network: one is used to eliminate none facial regions quickly, another for evaluating candidates carefully with fast multiple resolution technique. Szarvas et.al\cite{Szarvas2015Multi} used DCNNs for multi-view face detection named Deep Dense Face Detector(DDFD) which is a single model with last heat map for face classification and bounding box regression. Hongwei et.al\cite{Qin2016Joint} addressed previous cascade DCNNs structure\cite{Li2015A} which trained different stages isolate. Therefore, they jointly trained different stages to achieve better performance. Similar to \cite{Qin2016Joint}, Kaipeng et.al\cite{Zhang2016Joint} via DCNNs which is consist of three subnetwork for multiple task about face detection and alignment. These region-based and cascaded framework-integrated methodology forms a multi-scale input image pyramid or fix input size image and resize various size, it will perform several forward passes during inference and thus the computing consumption of the model will increase correspondingly.

Scale-variant methods: These methods will extract the feature and detect faces from various hierarchies in single network and then merge the predictions out from the various network hierarchies to generate the overall detection results. Following this designed style, Peiyun et al.\cite{Hu2016Finding} indicated that the context was helpful to detect tiny face in complicated scenes, and then they defined foveal descriptor to extract feature of tiny face in large receptive field. Inspired by \cite{Hu2016Finding}, \cite{Yang2017Face} proposed through scale-friendly DCNNs, which made use of training specialized networks with the most suitable depth and spatial pooling stride to detect face from each specific sub-range of scales. S.Zhang et al. \cite{Zhang2017S}put forward a single shot scale-invariant which uses VGG16 as backbone network and multiple feature hierarchies in single network for face detection. M.Najibi et al. \cite{Najibi2017SSH}designed a single stage headless network structure which was scale-invariant and could simultaneously detect faces with different scales from different layers in the way of single forward passing. 

\section{Our work}
\subsection{Fine-tuning SSD for face detection}
\ \ \ As the work\cite{Li2015A,Qin2016Joint,Zhang2016Joint}stated, their network component such as 12-net\cite{Li2015A}, branch x12\cite{Qin2016Joint}, or pnet\cite{Zhang2016Joint} employed region proposal mechanism to gain the ROI of the input image. For processing input image of multiple scales, these methods firstly construct pyramid for input image and then respectively input the image from pyramid into DCNNs to extract feature and detect face with region proposal mechanism. These framework whose feature pyramids built upon image pyramids required dense scale sampling such that it could achieve good results, at the same time it would cost more inference time and storage resources. In addition, multi-stage jointly training or testing increases the complexity of process flow. Inspired of scale-variant network designed style such as SSD which detects objects from various feature hierarchies in single network, the feature hierarchies have an inherent multi-scale, pyramidal hierarchical structure. Therefore, for reducing the complexity of training or testing face detector, we can directly use pyramidal feature hierarchy of DCNNs to detect face without constructing image pyramids. We try directly fine-tune SSD for face detection task. Some results are shown in Figure 1.\\
\begin{figure}[h!]
	\centering
	\subfloat[Detection result in WIDERFACE]{
		\label{fig:a}
		\begin{minipage}[t]{0.5\textwidth}
			\centering
			\includegraphics[scale=0.22]{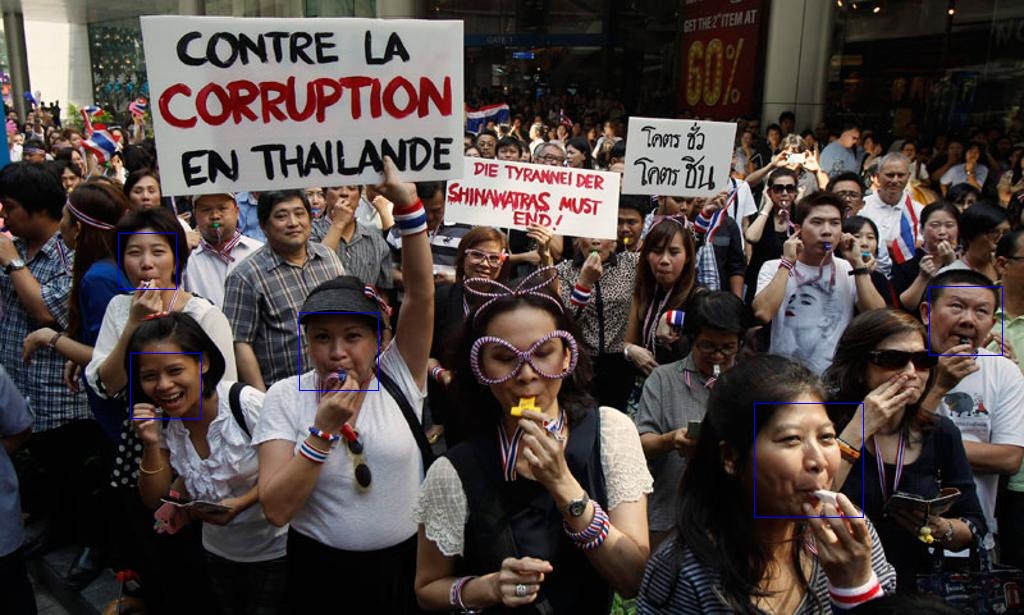}
		\end{minipage}
	}
	\subfloat[Detection result in FDDB]{
		\label{fig:b}
		\begin{minipage}[t]{0.5\textwidth}
			\centering
			\includegraphics[scale=0.4]{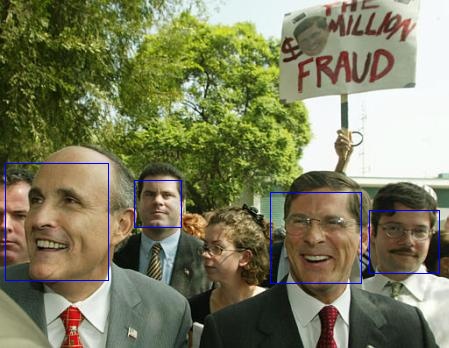}
		\end{minipage}
	}
	\captionsetup{font={scriptsize}}
	\caption{Some results detected by fine-tuning SSD}
\end{figure}

Although fine-tuned SSD demonstrated effective in face detection task as shown above, there exist some shortcomings in fine-tuned SSD, for instance, it has limit to detect smaller faces. Because DCNNs obtain a series of feature maps through forward propagating layer by layer with pooling operations. The feature hierarchy consisted of these feature maps has an inherent multi-scale, pyramidal shape. This network architecture style can reuse the multi-scale feature maps in the forward pass, and then add extra layers to build bottom-up feature pyramid for object detection of different scales. As the depth of network increasing, SSD will detect objects on lower-resolution maps of higher level layers in the feature hierarchy. However, it ignores the strong semantics which can improve detection performance and are computed at higher-resolution feature maps that are in the low level layers. Therefore, SSD-style using DCNNs' pyramidal feature hierarchy not only produces feature maps of different spatial resolutions, but also introduces large semantic gaps caused by different depths\cite{Lin2016Feature}. Consequently, It causes SSD limited for detecting small objects. \\
\begin{figure}[h!]
	\centering
	\subfloat[Distribution of WIDER FACE]{
		\label{fig:a}
		\begin{minipage}[t]{0.5\textwidth}
			\centering
			\includegraphics[scale=0.5]{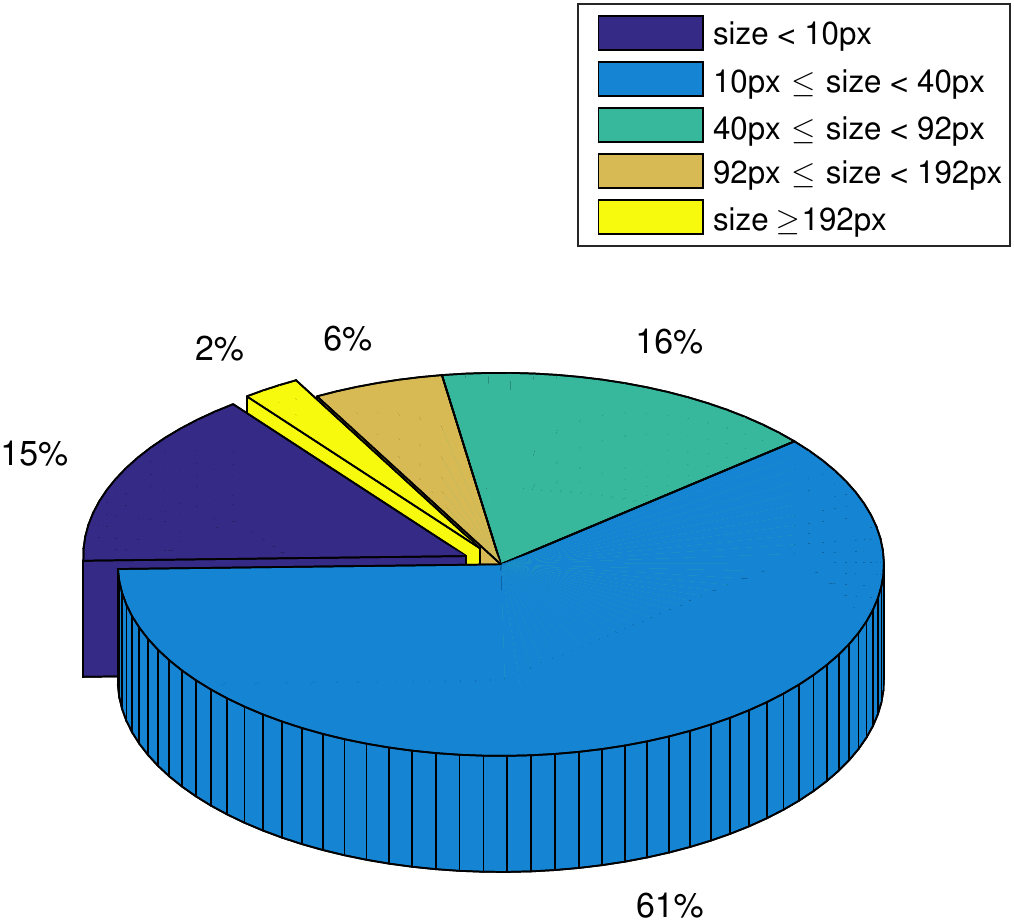}
		\end{minipage}
	}
	\subfloat[Distribution of AFLW]{
		\label{fig:b}
		\begin{minipage}[t]{0.5\textwidth}
			\centering
			\includegraphics[scale=0.5]{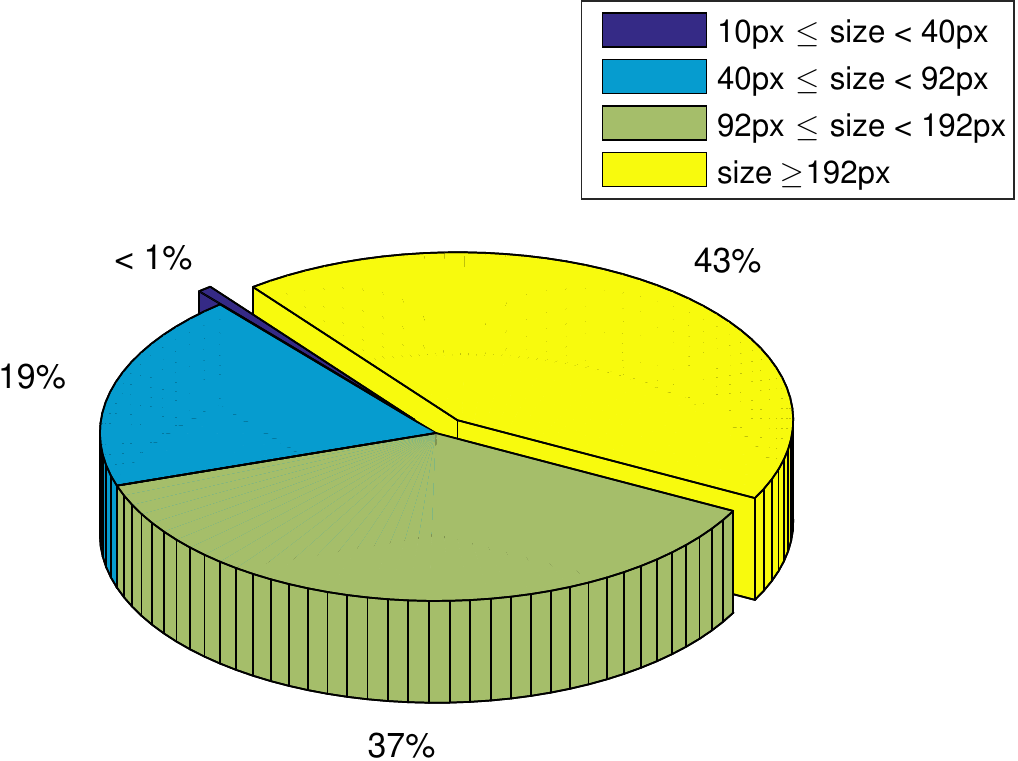}
		\end{minipage}
	}
	\captionsetup{font={scriptsize}}
	\caption{Distribution of face size in WIDER FACE and AFLW}
\end{figure}

We analyze the distributions of annotated faces in WIDER FACE\cite{Yang2015WIDER} and AFLW. As illustrated in Figure 2, we find most of annotated faces in WIDER FACE training datasets are very small and blurry, about 76\% faces whose size are less than 40px, and 15\% are less than 10px. In contrast, there are approximate 80\% faces whose size are more than 92px in AFLW, and 19\% are in the range of 40px-92px. Thence, for SSD this scale-variant without image pyramid and strong supervised learning methods, it is not suitable to detect small object using training set which is lack of ground truth one. Besides, there are approximate 15.1624\% faces in $1024\times{768}$ images, 17.0668\% ones in the range of $1024\times{682}-1024\times{684}$. Therefore, if we resize the original image to $300\times{300}$ which is as input size of SSD, the annotated small faces(<10px, or $\le10$px but < 40px) will get much smaller resulting in extracted feature lack of effective representative ability. At the same time, the detected bounding box of the receipt field located on that hierarchy is invariant. When this bounding box slides on that region, it could not discriminate the weak feature of small face from background. It leads to trained model has weak capacity to detect small face. 

\subsection{The designed network}
\ \ \ Our goal is to leverage the multiple scale-variant feature pyramid hierarchy for face detection in unconstrained scenes. Besides, as shown in figure 1, human can quickly detect small faces in other extreme and challenging conditions e.g. blurry, noised because we can utilize faces around semantic information such as hair, neck, or hat. Therefore we focus on how to comprehensively take advantage of strong semantic information in feature maps on different hierarchies extracted by the backbone network. We design a network likes encoder-decoder structure named Feature Hierarchy Encoder-Decoder Network(FHEDN) for face detection, which utilized hierarchy feature maps on different hierarchies extracted by the pre-trained backbone network e.g.VGG-16. The designed network consists of two key components named context modeling and prediction module. We will describe more details in following sections.

\subsubsection{Context modeling module}
\ \ \ \cite{Hu2016Finding} demonstrated the effect of context semantic information for detecting extreme small, blurry and noised face. How to modeling context to obtain strong semantic information that can improve the performance is the key technology of the whole work. Previous work e.g.\cite{Lin2016Feature,Fu2017DSSD,Kong2017RON,Cai2016A,Jonathan2017Fully} made use of subsequent deeper feature and upsampling procedure to refine context semantic information. Inspired by these work, we follow an hourglass like architecture called "encoder-decoder"\cite{Noh2015Learning,Newell2016Stacked} which encodes the feature embedding context semantical information in the medium layers, and then decodes the feature for special task in higher layers. This style of network is able to combine low-resolution, semantically strong features on low hierarchy with high-resolution, semantically weak features on high hierarchy through top-down pathway. Finally, it will produce a feature pyramid involving rich semantic information. Thence we adopt the paradigm of encoder-decoder to modeling context. The whole context modeling procedure contains deconvolution and element-wise sum operations. The deconvolutional layer is responsible to upsample current feature hierarchy, while element-wise sum layer is used to fuse feature on different layer. The context modeling module is shown in Figure 3.
	\begin{figure}[h!]
		\subfloat[Modeling mode A]{
			\label{fig:a}
			\begin{minipage}[t]{1\textwidth}
				\centering
				\includegraphics[scale=0.48]{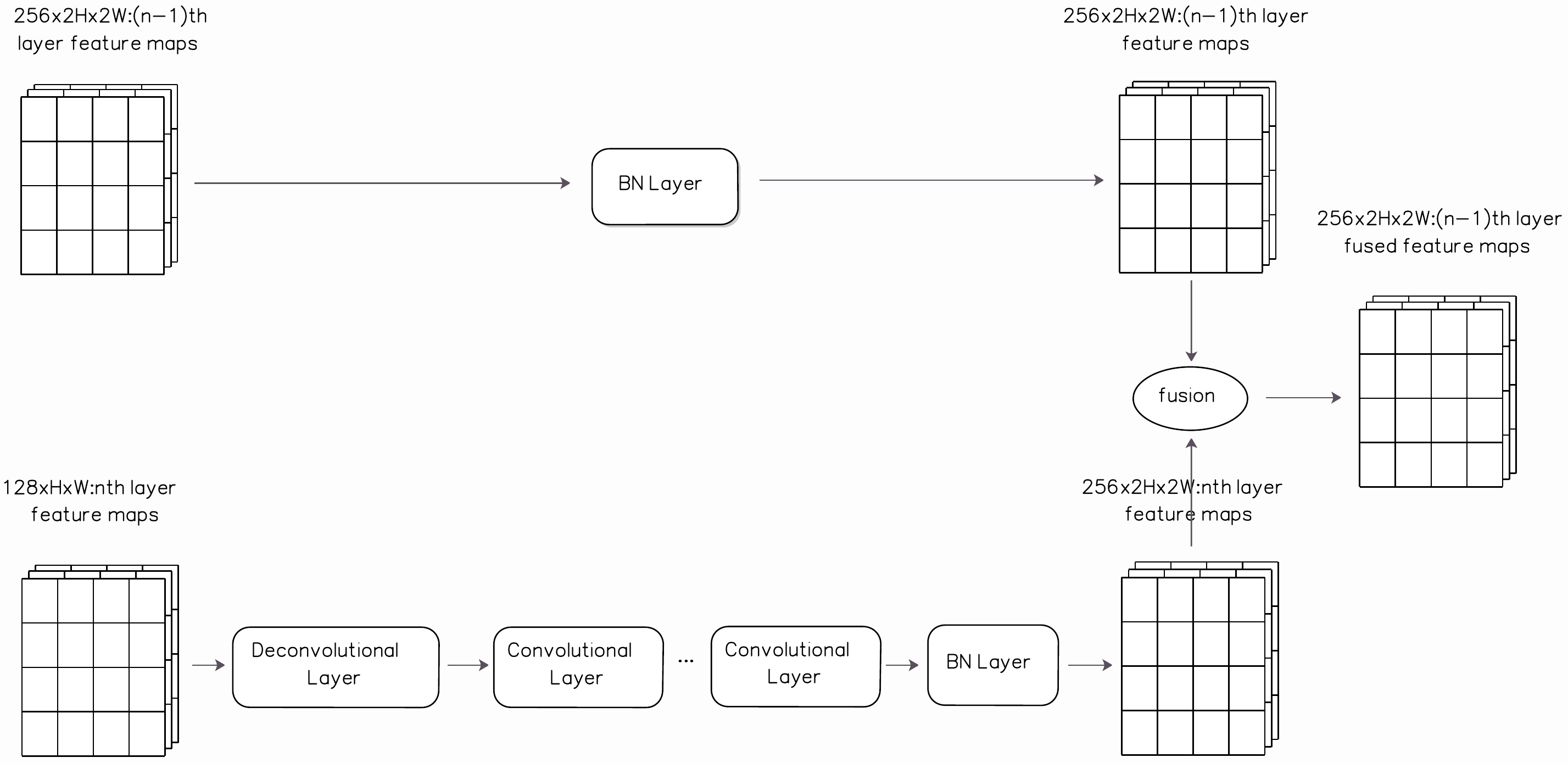}
			\end{minipage} 
		}\\
		\subfloat[Modeling mode B]{
			\label{fig:b}
			\begin{minipage}[t]{1\textwidth}
				\centering
				\includegraphics[scale=0.55]{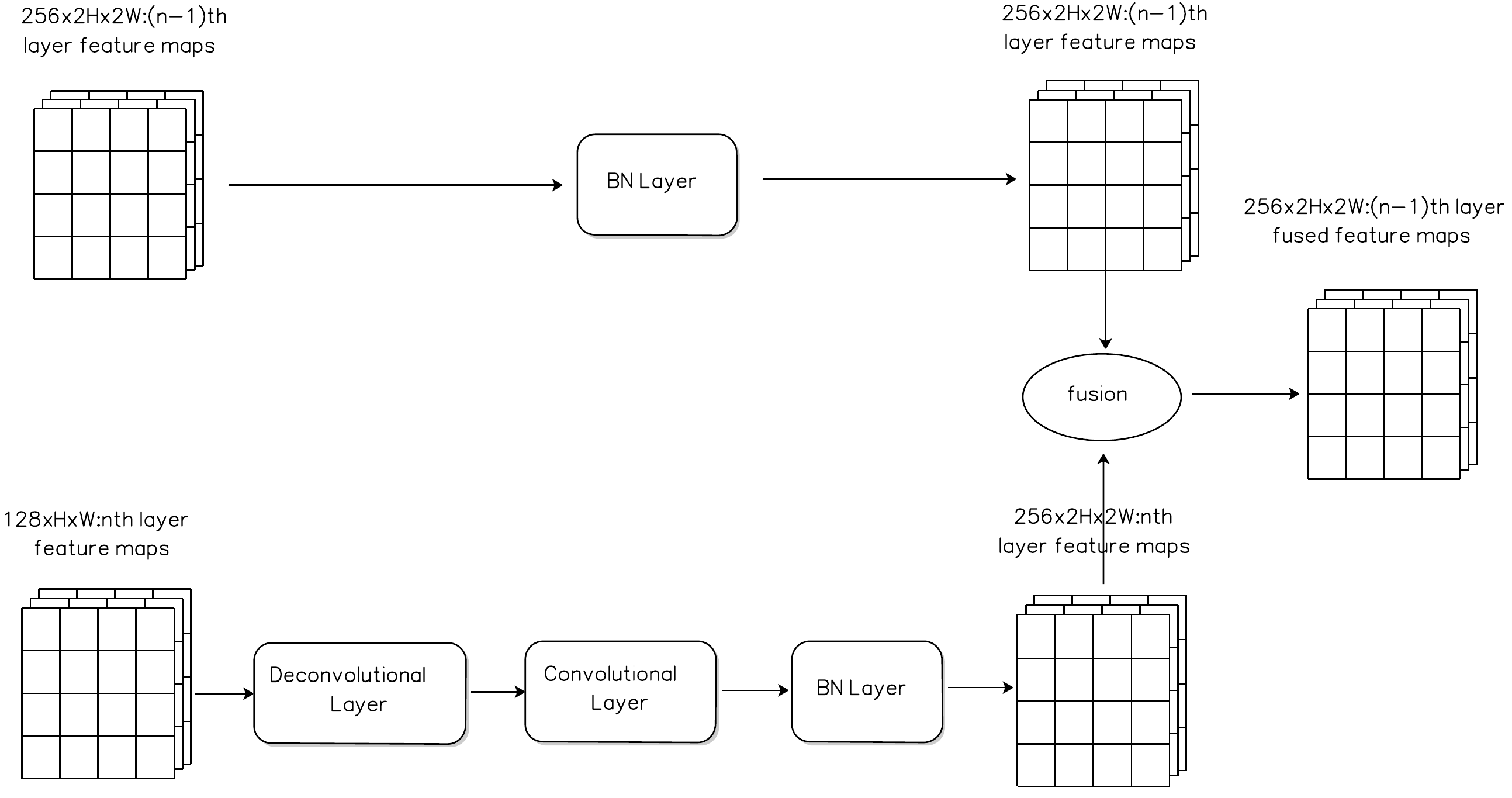}
			\end{minipage}
		}
	\captionsetup{font={scriptsize}}
	\caption{The context modeling module}
	\end{figure}

Here we try two type modeling modes as shown in Figure 3, we found mode A which contains three convolutional layers is not very helpful for improving performance. For reducing calculation redundancy, we　employ the other mode which only contains one convolutional layer to adjust channels fit for fusing. It not only works well but also does not sacrifice accuracy. 

\subsubsection{Prediction module}
\ \ \ Following the scale-variant feature hierarchy of designed style in SSD, a set of convolutional layer blocks as prediction modules are added after previous each scale hierarchy respectively. As shown in Figure 4, each prediction module is consist of two convolutional layers and default box generational layer. One convolutional layer is prepared for next softmax layer to classify face. The other one is provided to location regression. Default box generational layer is used to compute default boxes for corresponding feature hierarchy. The prediction module will produce a group of vectors, which is constructed by five value(first one represents confidence of face/background, other four represent the coordinate of left-top and right-bottom of detection box as respectively illustrated as white and gray cell in Figure 4).
\begin{figure}[h!]
		\centering
		\includegraphics[scale=0.5]{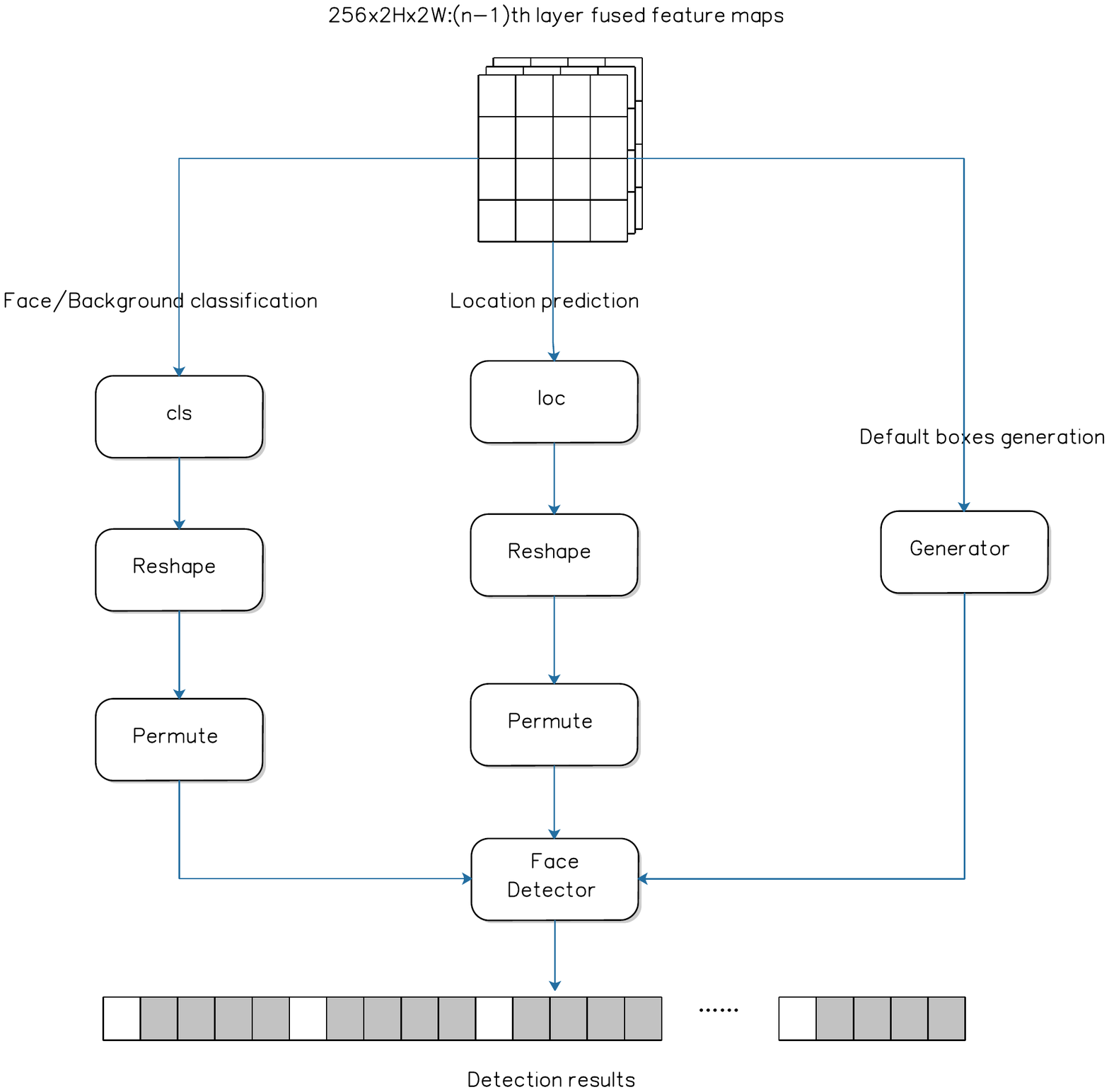}
		\captionsetup{font={scriptsize}}
		\caption{Prediction module}
\end{figure}

Suppose we want to compute the information of default box in some feature hierarchy. The calculation formula is 
\begin{equation}
	\begin{split}
		& cx = (x + \delta)\times{S_w},cy = (y + \delta)\times{S_h},w\in[0,w_f],h\in[0,h_f]\\
		& S_w = \frac{w_i}{w_f},S_h = \frac{h_i}{h_f}\\
		& w = s_k,h = w\times{a_r}
	\end{split}
\end{equation}
where $cx$ and $cy$ mean the center coordinate of the default box, $x,y$ represents index of the x and y axis on feature map in this hierarchy respectively. $w_f,h_f$ means the width and height of feature map respectively. $\delta$ is offset of the center coordinate of current default box to next one and is set 0.5. $S_w,S_h$ indicates stride of the default box on original detected image following x and y axis respectively and $h_i,w_i$ are width and height of the image. $w,h$ note the width and height of default box. $s_k$ denotes the size of the default box on the $k$th(total l hierarchies) hierarchy, which is computed as 
\begin{equation}
s_k = (s_0 + \frac{s_l - s_0}{l-2})\times{D_{min}},k\in[0,l)
\end{equation}
where $D_{min}=\min(M,N)$ means the min dimension of input image. $s_0$ and $s_l$ are computed as $\frac{RF_0}{D_{min}}$ and $\frac{RF_l}{D_{min}}$, where $RF_0,RF_l$ are receptive field size of 0-th and l-th scale hierarchy respectively.
$a_r$ is the aspect ratio for the default box. Similar to SSD, we add a default box whose hierarchy is $\sqrt{s_ks_{k+1}}$ for aspect ratio of 1.

\subsubsection{Network architecture and Training objective}
\ \ \ \textbf{Network architecture}\ \ \ Taking VGG-16 as the backbone network for example, it has 13 convolutional layers and 3 fully-connected layers. We convert both FC6 and FC7 to convolutional layers and retain frontal 13 convolution layers. Similar to SSD, we add extra convlolution layers to extend VGG-16 to improve its representational ability. Therefore, the overall network architecture is a fully convolutional network. As is shown in Figure 5, the overall network contains encoder and decoder part. Encoder subnetwork is used to extract deep feature from input image with hierarchies by hierarchies. Decoder subnetwork to integrate above described context modeling and prediction modules.
\begin{figure}[h!]
	\centering
	\includegraphics[scale=0.45]{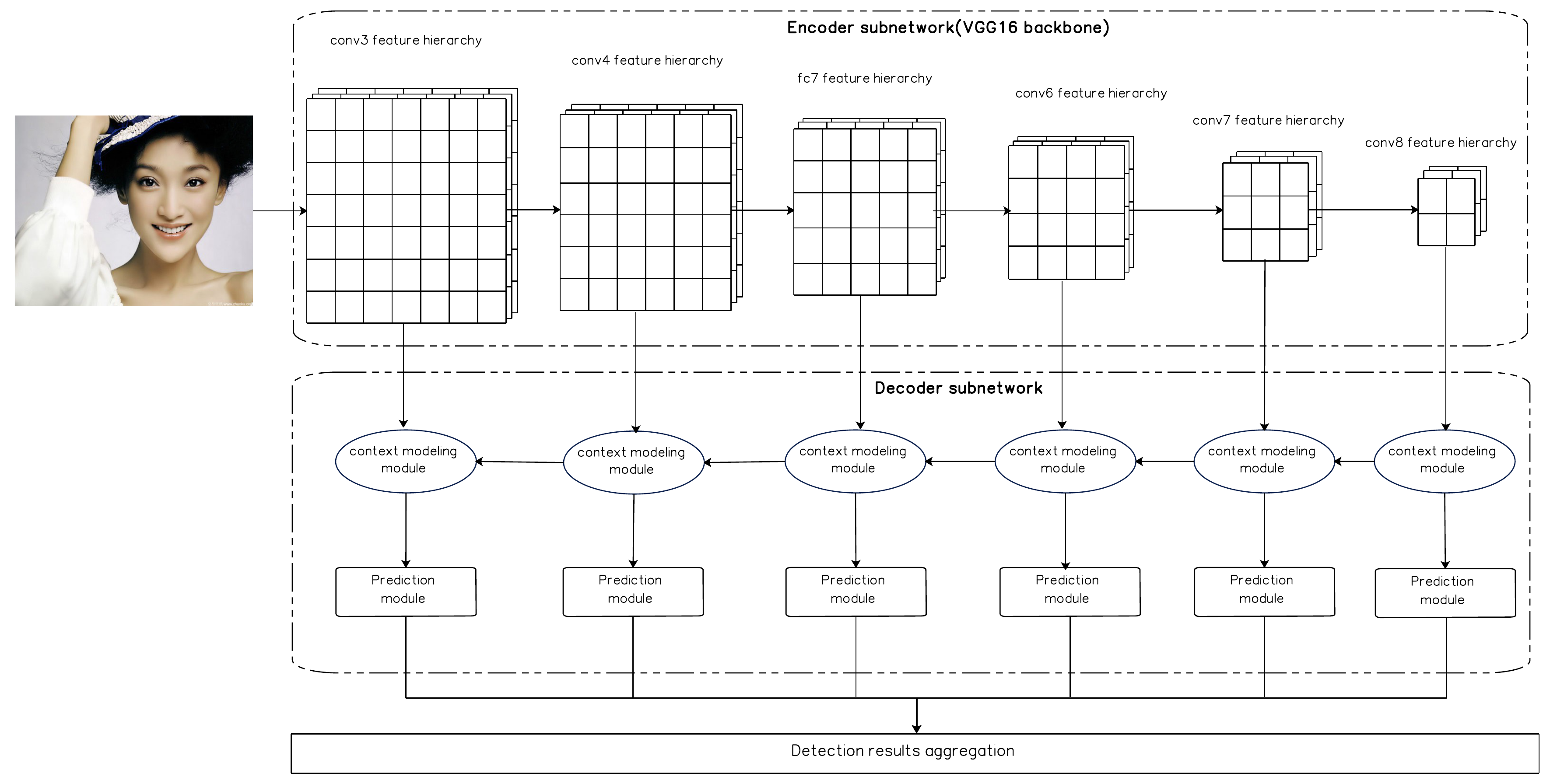}
	\caption{The whole network architecture}
\end{figure}

\textbf{Training objective}\ \ \ We employ the multi-task loss defined in \cite{Liu2015SSD} to jointly train our network: 
\begin{equation}
L(x,c,l,g)=\frac{1}{N}(L_{conf}(x,c)+\alpha L_{loc}(x,l,g))
\end{equation}
where N is the number of matched default boxes, $\alpha$ is used to balance above two loss terms and is set to 1, $x_{ij}={1,0}\in x$ indicates the i-th default bounding box matching to the j-th annotated face box, $c$ is the confidence of face or background, $l$ and $g$ respectively notes the predicted box and the annotated face box. We first match each face to the default box with the maximum jaccard overlap. When the predicted bounding box has the jaccard overlap greater than a threshold, it will be assigned to a face.

$L_{conf}(x,c)$ represents category confidence loss, which is adopt 2-class softmax loss function for face detection:
\begin{equation}
	\begin{split}
		& L_{conf}(x,c)=-\sum_{i\in Pos}^Nx_{ij}\log(\hat{c}_i)-\sum_{i\in Neg}\log(\hat{c}_i^0)\\
		& \hat{c}_i=\frac{\exp(c_i)}{\sum_p\exp(c_i)}
	\end{split}
\end{equation}
where $\hat{c}_i$ notes the predicted confidence of face about i-th default box, while $\hat{c}_i^0(\approx1-\hat{c}_i)$ notes the predicted confidence of background box about i-th default box.

$L_{loc}(x,l,g)$ is a smooth L1 loss function\cite{Girshick2015Fast}, whose value denotes loss between the predicted box(l) and the ground truth box(g). Following by \cite{Liu2015SSD}, we use it to regress to offsets between default box and annotated face.
\begin{equation}
	\begin{split}
		& L_{loc}(x,l,g)=\sum_{i\in Pos}^N\sum_{m\in{cx,cy,w,h}}x_{ij}smooth_{L1}(l_i^m-\hat{g}_j^m)\\
		& \hat{g}_j^{cx}=(g_j^{cx}-d_i^{cx})/d_i^w,\hat{g}_j^{cy}=(g_j^{cy}-d_i^{cy})/d_i^h\\
		& \hat{g}_j^{w}=\log\frac{g_j^w}{d_i^w},\hat{g}_j^h=\log\frac{g_j^h}{d_i^h}
\end{split}
\end{equation}
where smooth L1 loss function defined in \cite{Girshick2015Fast} is
\begin{equation}
	smooth_{L1}(x)=\left\{
	\begin{split}
		& 0.5x^2 & \ if\ |x|<1\\
		& |x|-0.5 & \ otherwise
	\end{split}
	\right.
\end{equation}

\section{Experiments}
\subsection{Experiment settings}
\ \ \ \textbf{Training datasets}\ \ \ Firstly, we use AFLW to fine-tune original SSD for face detection task. AFLW contains 25,993 annotated faces in real-world images. We choose 18,303 correctly annotated faces of images from AFLW, and then randomly select 80\% of those as training dataset, while the remaining is reserved as validation dataset. As section 3.1 stated, because of distribution of AFLW and weak capability of SSD for detecting small object, fine-tuned SSD dose not have sufficient capability to detect face in extreme conditions. Secondly, we consider another dataset WIDER FACE which contains 32,203 images and 393,703 annotated faces and has more than 50\% extreme small annotated faces. The dataset is split into training(40\%), validation(10\%) and testing(50\%). It is very suitable for scale-variant without feature pyramid to train face detector on extreme scenes. 
 
\textbf{Testing datasets}\ \ \ For evaluating the effectiveness of our network, we verify FHEDN on two public face detection benchmarks. One is Face Detection Data Set and Benchmark(FDDB)\cite{Jain2010FDDB}, which contains 2,845 images with a total of 5,171 annotations including occlusions, difficult poses and low resolutions. The other is WIDER FACE validation and test set. We follow the standard evaluation protocol on FDDB using receiver operating characteristic curve(ROC) with two metric: discontinuous score and continuous score. While for WIDER FACE validation and test set, we use average precision(AP) as evaluation metric.

\textbf{Experiment platform}\ \ \ We implement our experiments on Caffe framework\cite{Jia2014Caffe}. At the same time, we apply and modify some source code provided by \cite{Liu2015SSD} in order to be suitable for our task. Besides, we try various hyper parameters for training with Stochastic Gradient Descent(SGD) algorithm. The size of each input image in a batch is set $512\times{512}$. The network is trained on NVIDIA Tesla P40 leased in cloud computing server with a total of 14 images per mini-batch. Weight decay is 0.00001 and momentum is 0.9. Meanwhile, the initial learning rate is set 0.01 and it will be dropped by 10 at 40480 and again at 70000 iterators with total 80000 iterators. 

\subsection{Implement detail}
\subsubsection{Influence of scale}
\ \ \ In this section, we analyze the influence of scale of feature maps in different hierarchies to the detected performance. Taking VGG16 backbone network as example, we show the output feature maps which will be fuse with previous hierarchy to modeling context in Figure 6. As illustrated in Figure 6(a), the output feature maps of conv3\_3, conv4\_3, fc7 were computed without normalization operation, whose scale range are different from conv6\_2 and conv7\_2. It will cause the feature map in large scale range covers the one in low scale range when it is in feature fusing stage. Different from \cite{Liu2015ParseNet} used L2 normalization technique, we utilize batch normalization layer attached above layers before fusing to solve this problem. Parameters in batch normalization layer are trained by network without manual setting. Figure 6(c) and (d) show the training loss curves about the influence of normalization. It demonstrates normalization operation could keep the training process more stable. 
\begin{center}
	\begin{figure}[h!]
		\centering
		\subfloat[Before normalized histogram of feature maps]{
			\label{fig:a}
			\begin{minipage}[t]{0.5\textwidth}
				\includegraphics[scale=0.45]{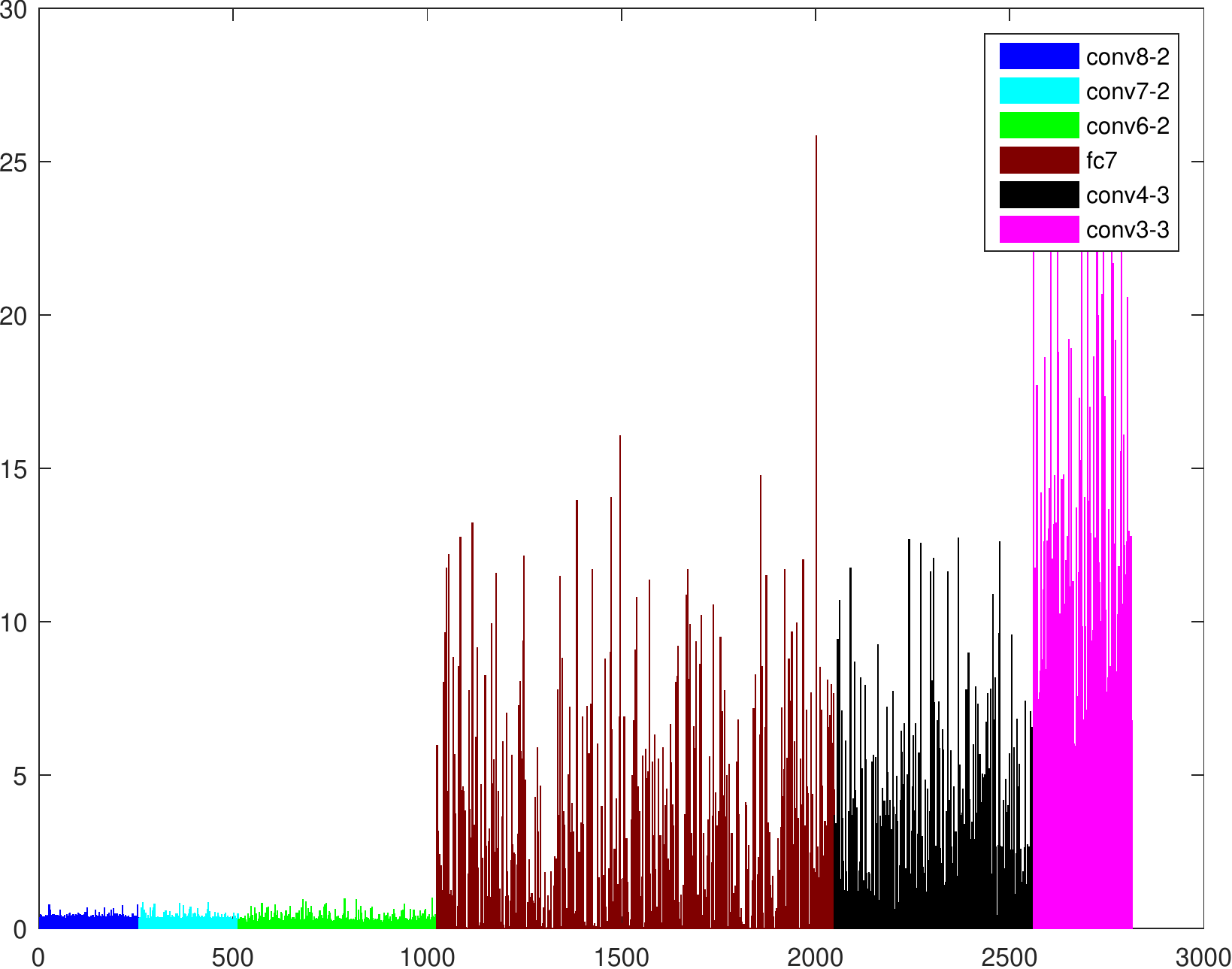}
			\end{minipage}
		}
		\subfloat[After normalized histogram of feature maps]{
			\label{fig:b}
			\begin{minipage}[t]{0.5\textwidth}
				\includegraphics[scale=0.45]{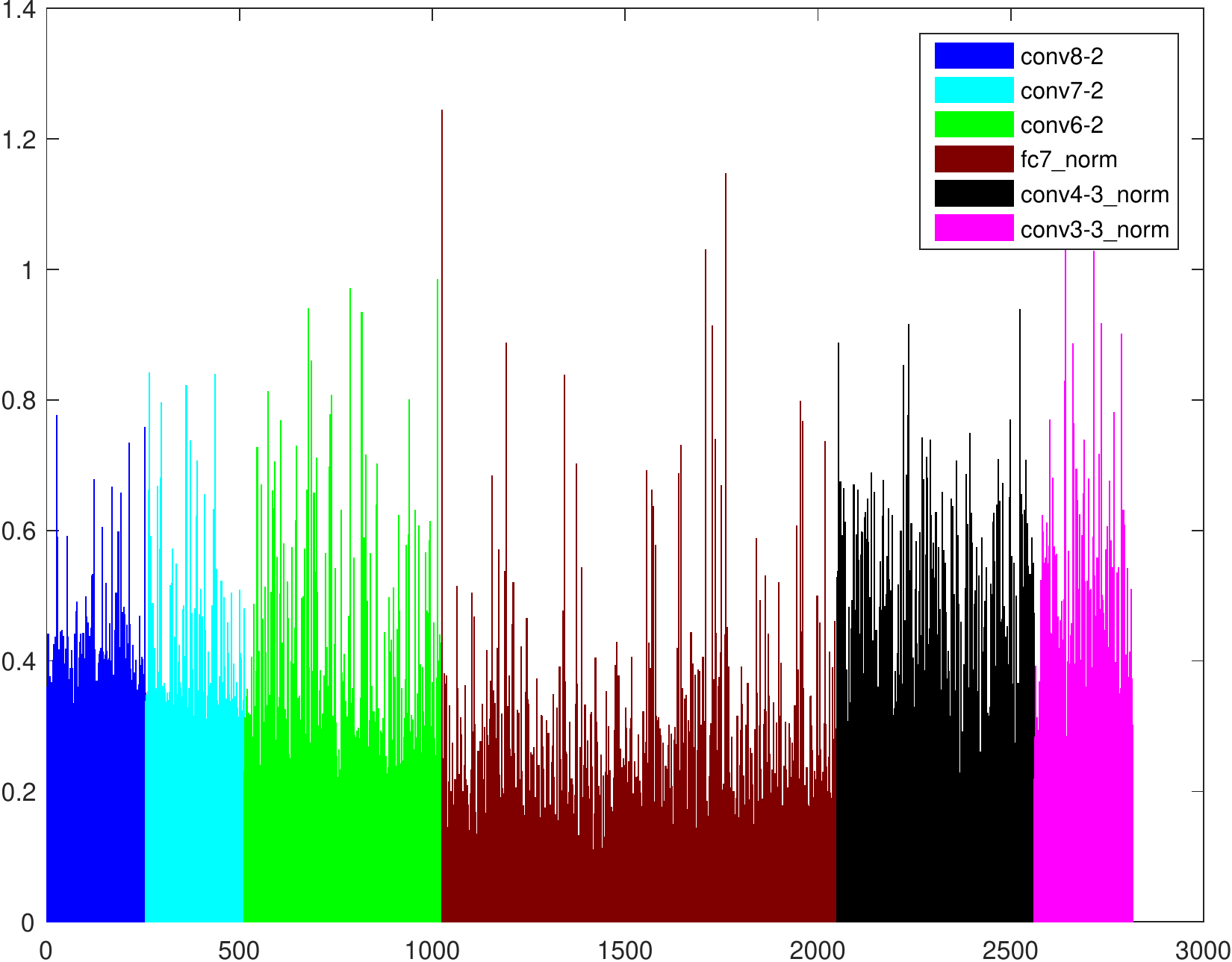}
			\end{minipage}
		}\\
		\centering
		\subfloat[Without normalized training loss curve]{
			\label{fig:c}
			\begin{minipage}[t]{0.5\textwidth}
				\includegraphics[scale=0.6]{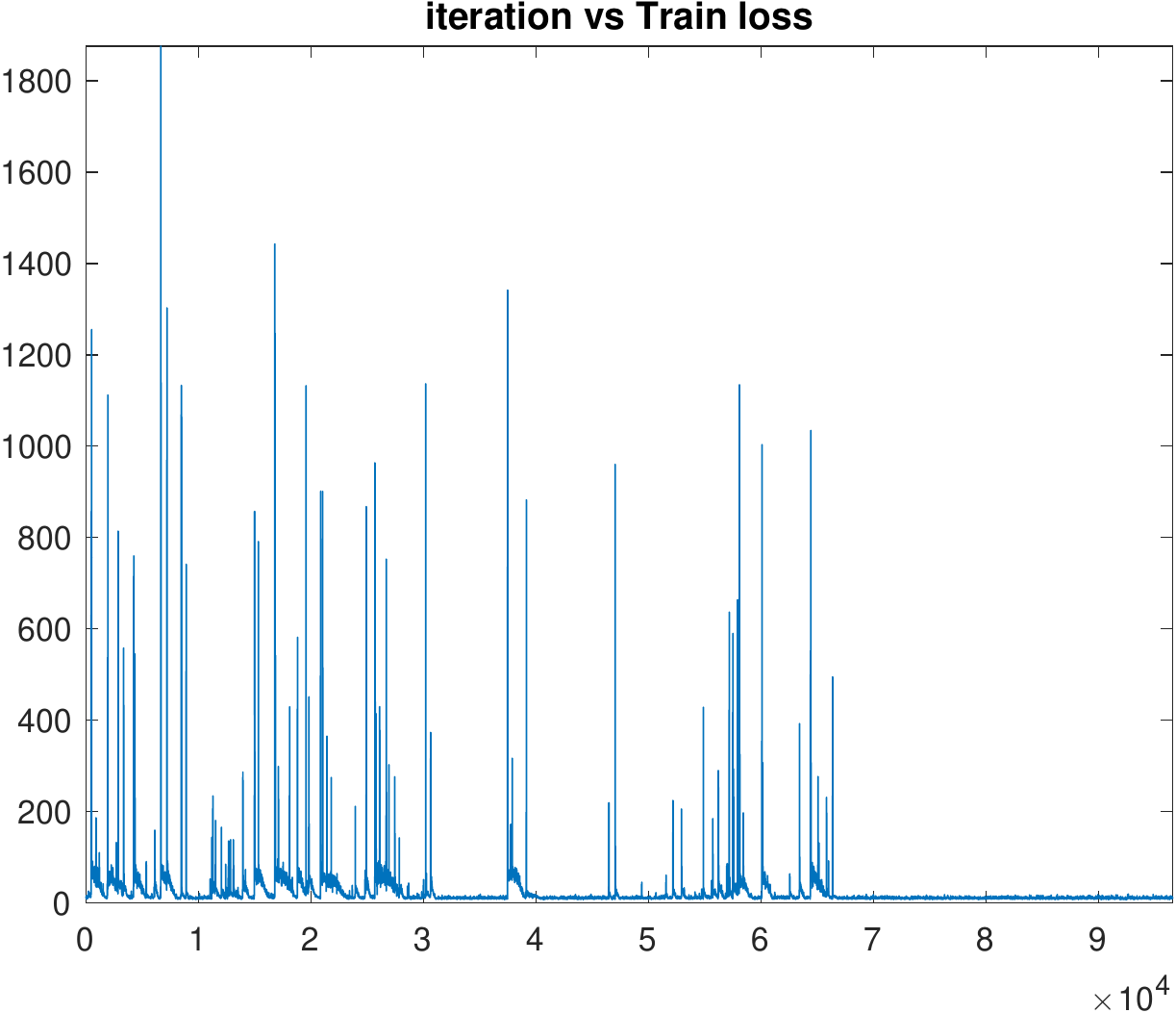}
			\end{minipage}
		}
		\subfloat[With normalized training loss curve]{
			\label{fig:d}
			\begin{minipage}[t]{0.5\textwidth}
				\includegraphics[scale=0.6]{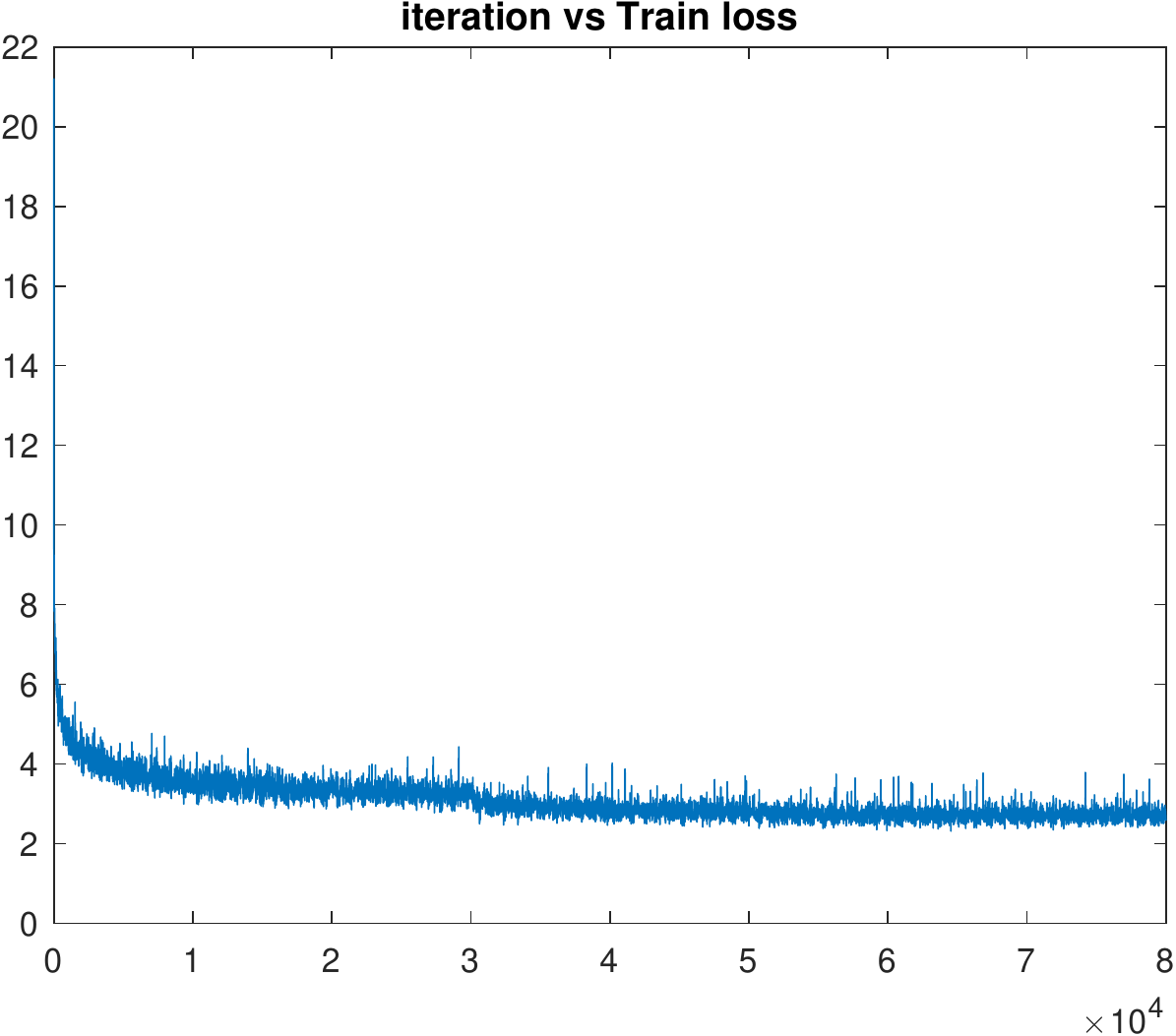}
			\end{minipage}
		}
		\captionsetup{font={scriptsize}}
		\caption{The influence of normalization}
	\end{figure}
\end{center}

\subsubsection{Influence of receptive field}
\ \ \ The receptive field(RF) size is a crucial issue in computer vision because it need be large enough to capture information about the covered region. Theoretical receptive field(TRF) can be computed from designed network architecture. Zhou et al.\cite{Bolei2014Object} introduce the concept of empirical receptive field via a data-driven approach to demonstrate that the actual size of RF is much smaller than TRF. Luo et al.\cite{Luo2017Understanding} adopt mathematical model to analyze the empirical receptive field relate to TRF and put forward effective receptive field, who pointed out only a fraction of TRF would contribute equally to an output unit's response. As stated in section 3.2.2, the prediction module contains several components, one is the key part to generate default boxes. If the size of generated default box is not suitable for the receptive field, the extracted feature on corresponding receptive field do not represent the default box. It will make trained model behave weak representable capacity. Wei et.al\cite{Xiang2017Context} analyzed effective receptive field within the framework of object detection and provide the algorithm to calculate the effective receptive field sizes of a standard VGG16 network. We use \cite{Xiang2017Context} provided algorithm to design default box to solve above described size problem. Take an example as eltw13 layer to detect small faces, we use formula (1) to compute the default box size $\in[10.24,30.72]$, whose range can cover the effective receptive field size of conv3\_3.

In addition, all the image in WIDER FACE has the same width 1024px but various height. When we resize the original image to $512\times{512}$ for training, the original aspect ratio of annotated face will change. Therefore, the default box should be adjusted to reduce the affect of distortion by resizing operation. We count top3 percent of face size about original image size in WIDER FACE training dataset as Table 1. For example, images which contain face size is in $\in(0,10]$ are approximately 1932  ($12880\times{15\%}$, as illustrated in Figure 1(a)), 22.1640\% of them are in $768\times{1024}$. We find face size mainly concentrated in $768\times{1024}$, $683\times{1024}$ from Table 1 listed. We compute the aspect ratio in responding resized image is approximately 1.5. Analogously, for images in  $576\times{1024}$ and $1536\times{1024}$, we approximately estimate their new aspect ratio as 0.5. In the end, the overall new aspect ratio is \{0.5,1,1.5,$\frac{1}{0.5}$,$\frac{1}{1.5}$\}.  

\begin{table}
	\centering
	\caption{The top 3 percent of size in the corresponding range}  
	\begin{tabular*}{14cm}{llll}  
		\hline  
			face size(px) & top1 & top2 & top3\\  
		\hline  
			0<size<10 & $768\times{1024}$(22.1640) & $683\times{1024}$(8.5002) & $576\times{1024}$(6.9691)\\
		\hline 
			10$\le$size<40 & $768\times{1024}$(14.7147) & $683\times{1024}$(10.1768) & $684\times{1024}$(4.1434)\\ 
		\hline
			40$\le$size<92 & $768\times{1024}$(13.4199) & $683\times{1024}$(11.4456) & $682\times{1024}$(4.2862)\\
		\hline    
			92$\le$size<192 & $768\times{1024}$(10.4288) & $683\times{1024}$(8.1915) & $1024\times{1024}$(3.8452) \\
		\hline
			192$\le$size& $768\times{1024}$(5.7435) & $1536\times{1024}$(4.9042) & $683\times{1024}$(4.7731)\\  
		\hline
	\end{tabular*}  
\end{table} 

\subsubsection{Other implementation details}
\textbf{Data augmentation}\ \ \ We apply data augmentation to make the proposed network more robust to various input sizes and shapes. Each entire original input image is randomly sampled a patch, whose cropped ratio is select from 0.3, 0.5, 0.7, 0.9 or 1.0. Then, we set the minimum jaccard overlap with the face is 0.5, 0.7 or 0.9, which can be help to extract feature of some occluded faces. After random cropping, the sampled patch will be horizontally flipped with probability of 0.5 and applied some photo-metric distortions.\\
\textbf{Online hard example mining}\ \ \ Since the number of negative default box is more than the positive, there is a significant imbalance between the positive and negative training examples. For stable and faster optimization training, we apply online hard example mining(OHEM)\cite{Shrivastava2016Training} technique to resample hard examples during training stage. After OHEM, the positive default boxes with the lowest scores and the negative default boxes with highest scores are randomly selected so that the ratio between the negatives and positives is at most 3:1. 

\subsection{Results}
\ \ \ We evaluate our network against state-of-the-art face detection methods on two benchmark datasets: FDDB and WIDER FACE.

\subsubsection*{FDDB results}
\ \ \ We divide the FDDB into 10 folds for performance evaluation and accumulate the detection results to generate the Receiver Operating Characteristic(ROC) curves. Besides, because of ellipse region style adopted in FDDB and for a more fair comparison under the continuous score evaluation, we transform the predicted bounding boxes to meet FDDB annotation style following the toolbox provided by \cite{Jain2010FDDB}. The results compared with other state-of-the-art\cite{Li2015A,Liao2016A,Szarvas2015Multi,Qin2016Joint,Zhang2016Joint,Hu2016Finding,Zhang2017S,Yang2017Face,Wang2017Detecting} are shown in Figure 7(a) and Figure 8(b). 
	\begin{figure}[h!]
		\centering
	\includegraphics[scale=0.4]{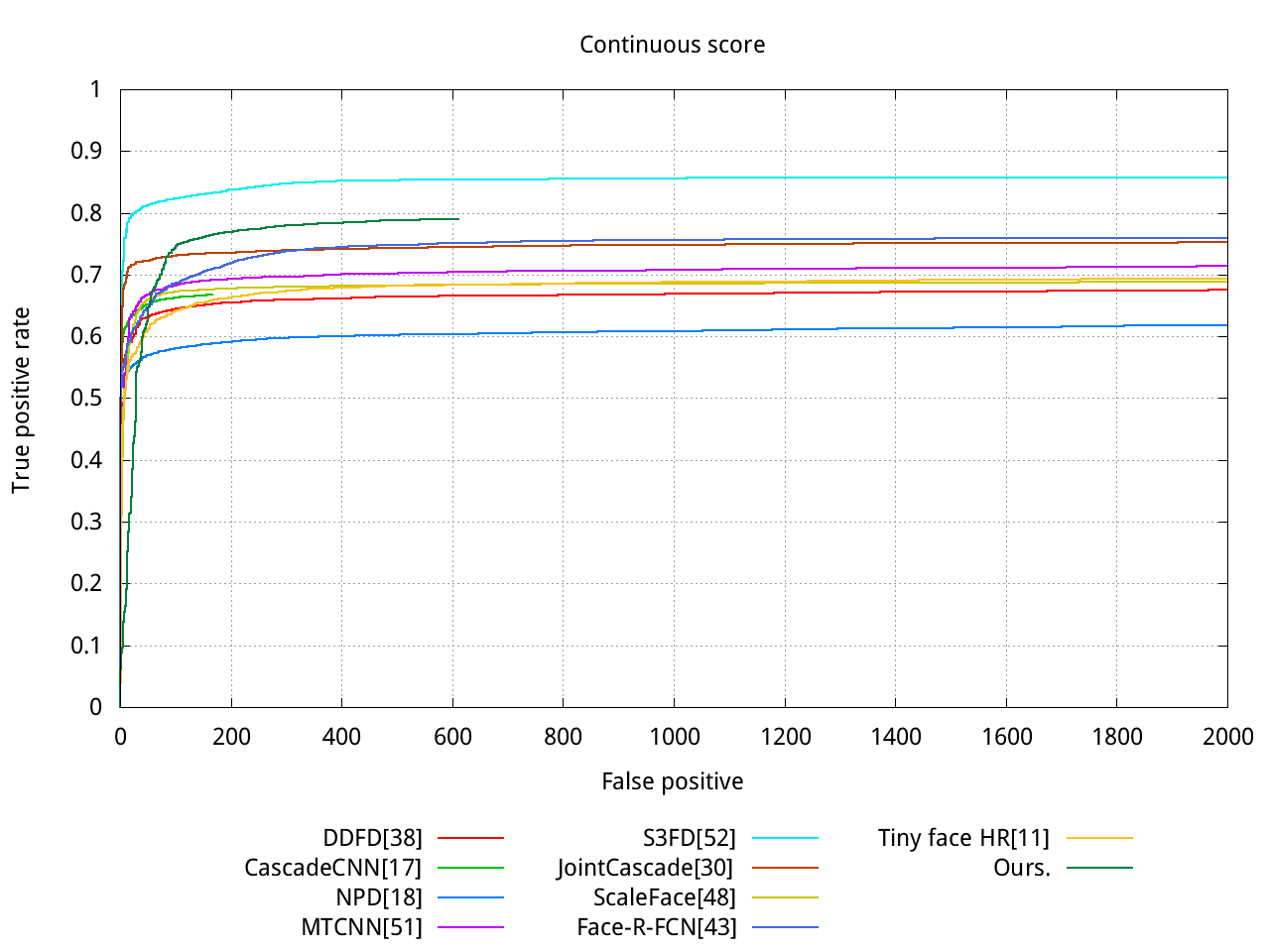}\\
	\includegraphics[scale=0.4]{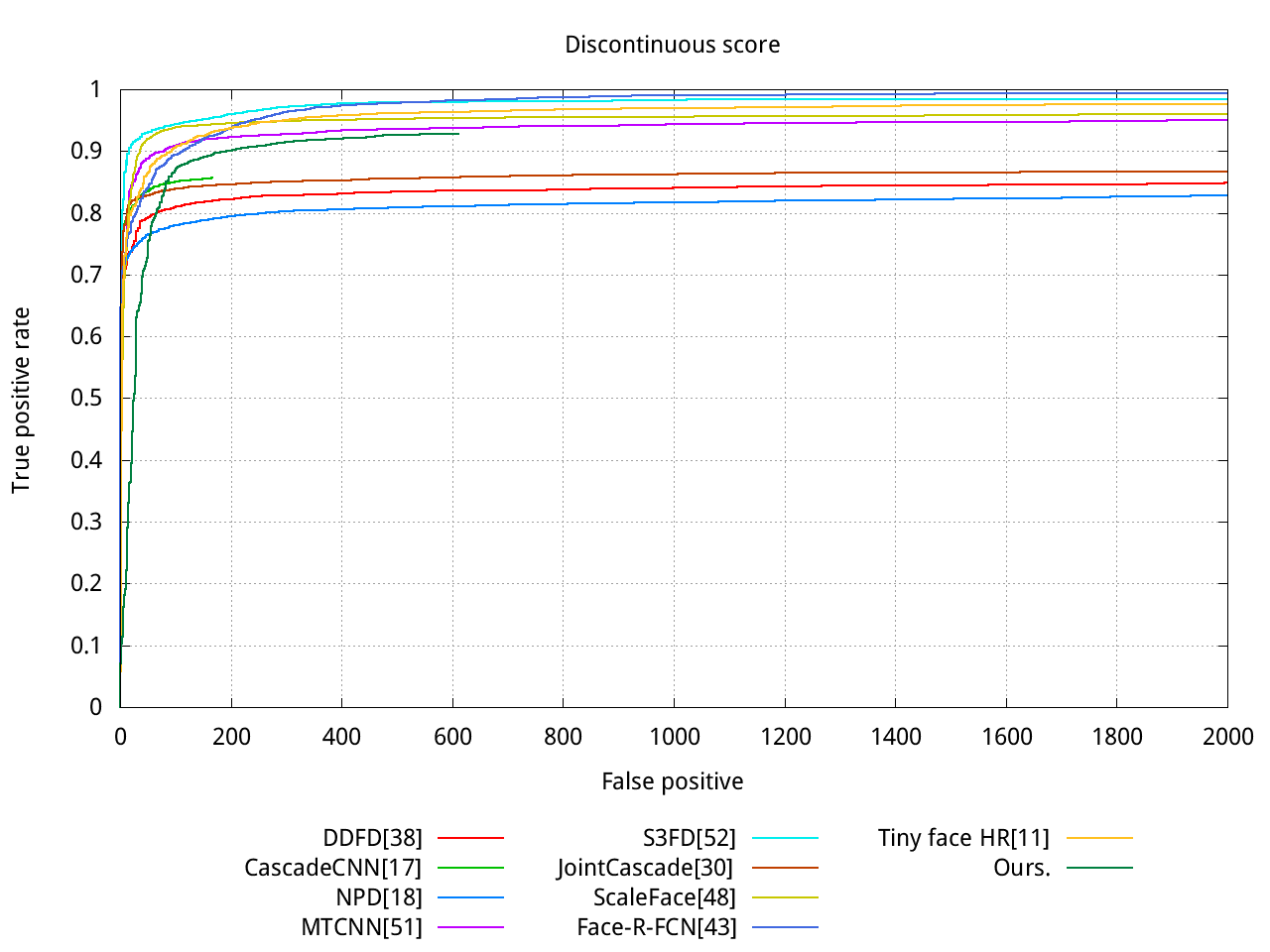}
	\centering
	\captionsetup{font={scriptsize}}
	\caption{Evaluation on FDDB benchmark}
\end{figure}

\subsubsection*{WIDER FACE results}
\ \ \ We evaluate and compare our method with top performances\cite{Yang2014Aggregate,Yang2016From,Zhang2016Joint,Ohn2017To,Zhu2016CMS,Hu2016Finding,Najibi2017SSH,Zhang2017S,Yang2017Face} on validation and test set split from WIDER FACE. Both these two benchmark datasets is divided into three levels(Easy, Medium and Hard subset) according to the difficulty settings correlate with the face scales. The precision-recall curves and mAP values are shown in Figure 7. These results demonstrate the effectiveness of the proposed method in detecting small and hard faces in unconstrained scenes. 
\begin{table}
	\centering
	\caption{Evaluation on WIDER FACE compared with the sate of the art}  
	\begin{tabular*}{8cm}{llll}  
		\hline  
		Alogrithms & Easy & Medium & Hard\\
		\hline
		ACF-WIDER\cite{Yang2014Aggregate} & 65.9\% & 54.1\% & 27.3\%\\  
		\hline
		Faceness\cite{Yang2016From} & 71.6\% & 60.4\% & 31.5\%\\  
		\hline  
		MTCNN\cite{Zhang2016Joint} & 85.1\% & 82.0\% & 60.7\%\\
		\hline 
		LDCF+\cite{Ohn2017To} & 79.0\% & 76.9\% & 52.3\% \\
		CMS-RCNN\cite{Zhu2016CMS} & 89.9\% & 87.4\% & 77.2\%\\ 
		\hline 
		HR\cite{Hu2016Finding} & 92.3\% & 91.0\% & 81.9\%\\ 
		\hline
		SSH\cite{Najibi2017SSH} & 93.1\% & 92.1\% & 84.5\% \\
		\hline    
		S3FD\cite{Zhang2017S} & 93.7\% & 92.5\% & 85.9\% \\
		\hline
		ScaleFace\cite{Yang2017Face} & 86.8\% & 86.7\% & 77.2\%\\
		\hline
		FHEDN(ours) & \textbf{87.1\%} & \textbf{83.1\%} & \textbf{63.4\%}\\  
		\hline
	\end{tabular*}  
\end{table} 

\begin{figure}[htbp]
	\centering
	\subfloat[Val: Easy]{
		\label{fig:a}
		\begin{minipage}[t]{0.3\textwidth}
			\includegraphics[scale=0.3]{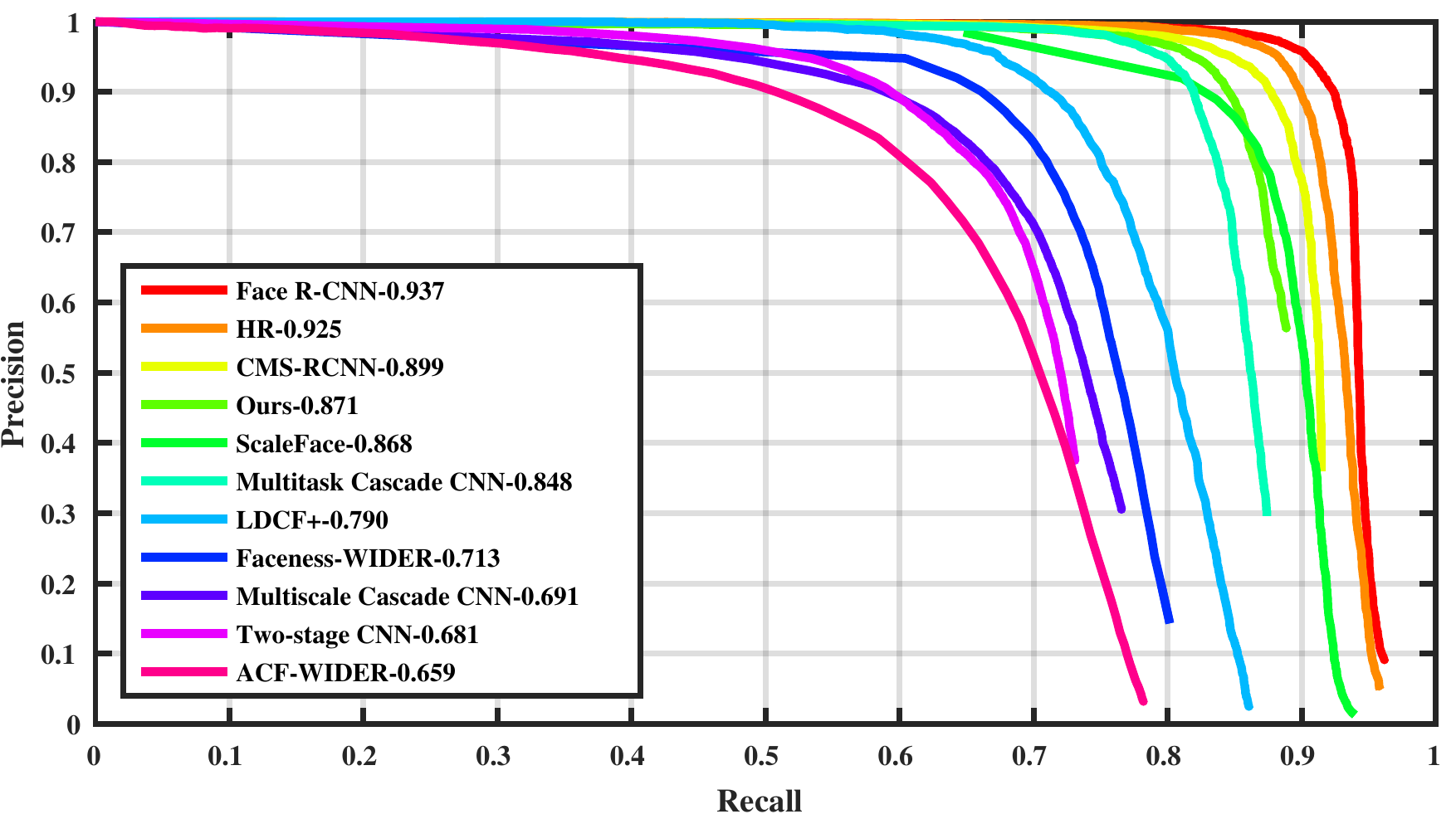}
		\end{minipage}
	} 
	\subfloat[Val: Medium]{
		\label{fig:b} 
		\begin{minipage}[t]{0.3\textwidth} 
			\includegraphics[scale=0.3]{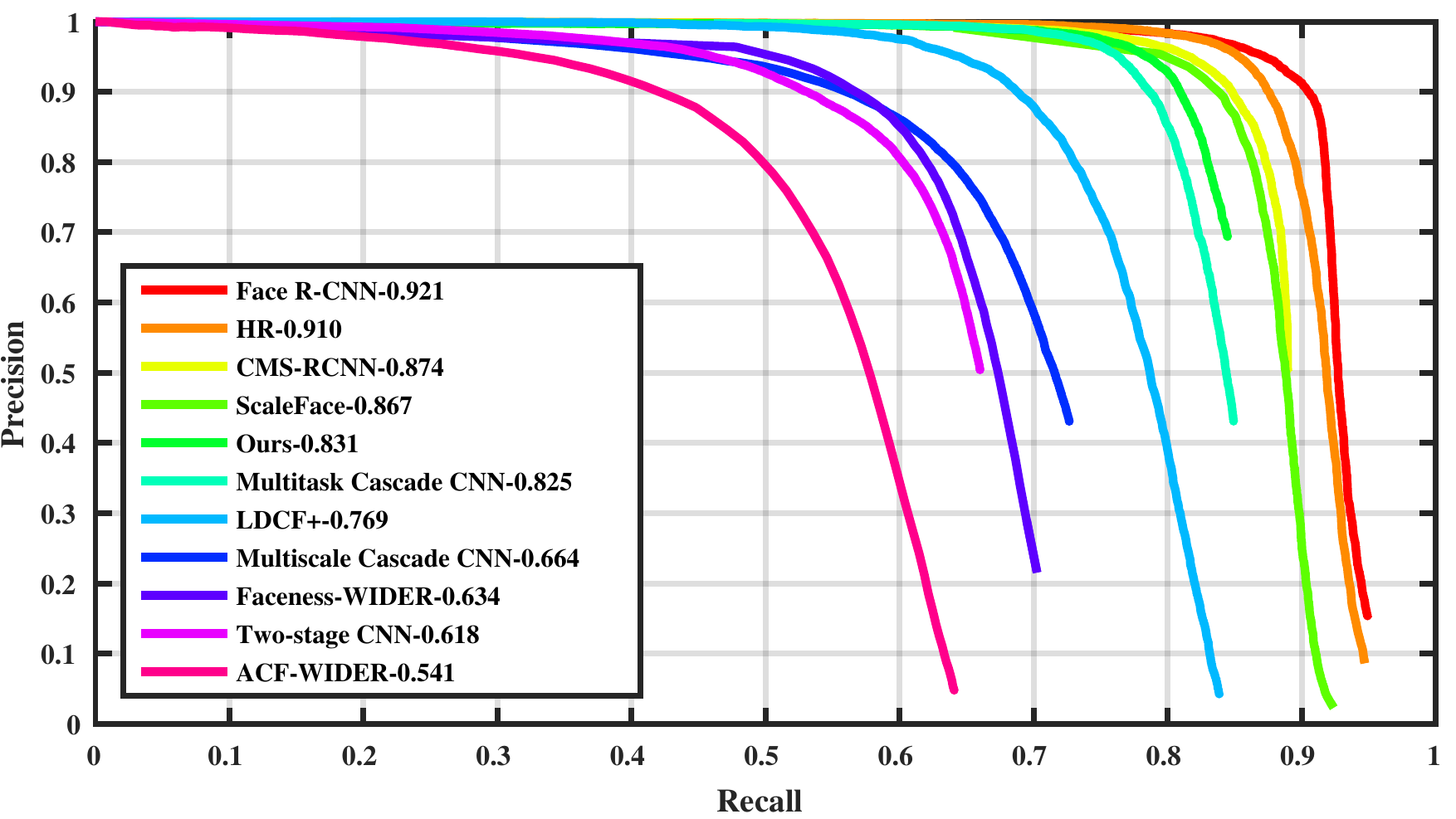} 
		\end{minipage}
	}  
	\subfloat[Val: Hard]{
		\label{fig:c}
		\begin{minipage}[t]{0.3\textwidth}
			\includegraphics[scale=0.3]{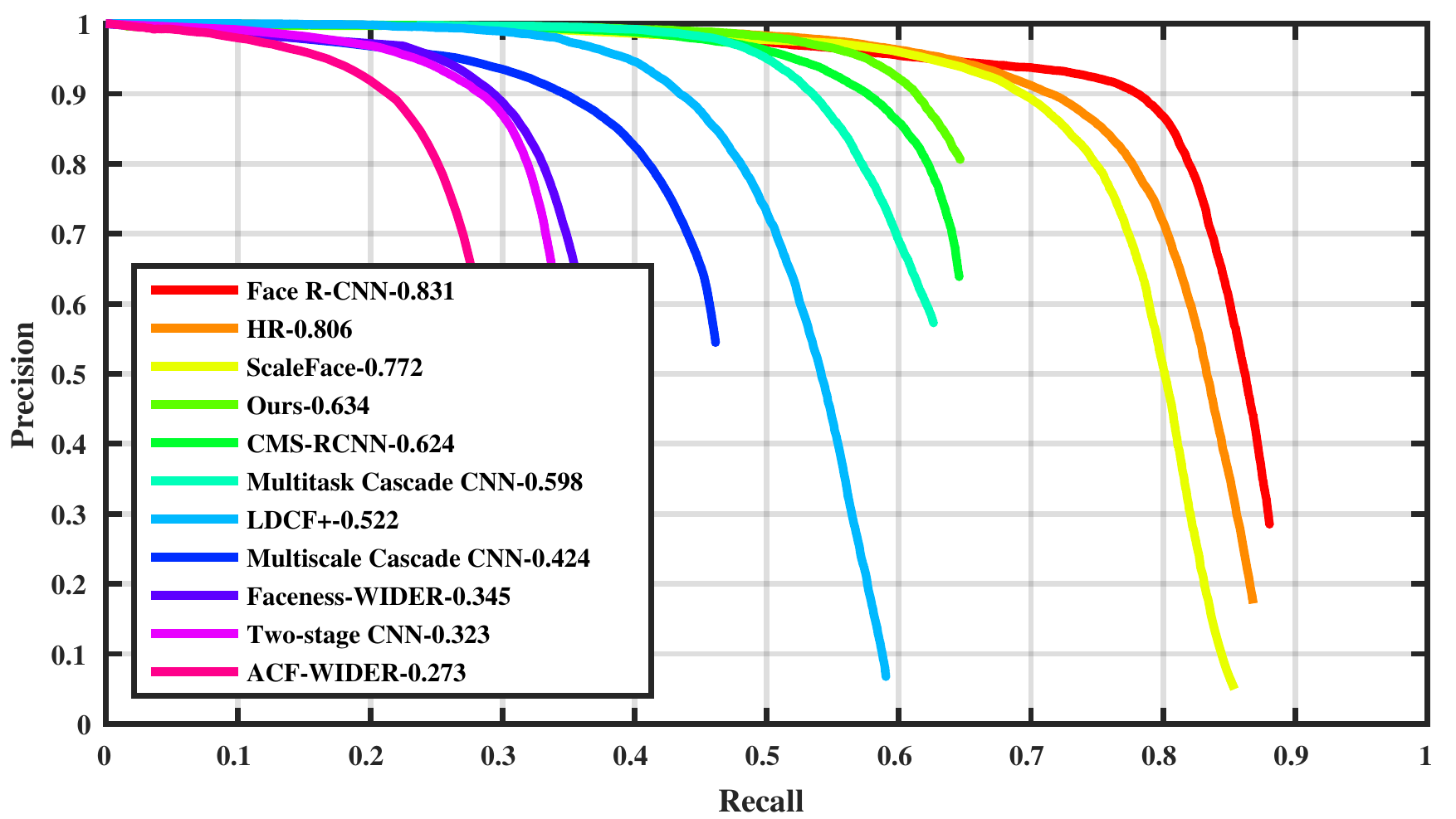}
		\end{minipage}
	}
	\captionsetup{font={scriptsize}}  
	\caption{Precision recall curves on WIDER FACE validation and test set}
\end{figure}

\section{Conclusion}
\ \ \ In this work, we have designed an end-to-end effective network with multiple scales feature hierarchy to detect faces in unconstrained scenes. Firstly, we fine-tune SSD as face detector via AFLW dataset and analyze its shortcoming for small, blur and occluded face detection. Secondly, we design a feature hierarchy network named FHEDN to improve detection performance, which fuse context semantic information fused with deeper feature hierarchies. Last but not least, we analyze some devil in implement details by statistical form and find some solutions to further improve the performance of our proposed method. Although there is a gap between our method and sate of the art ones, our designed network has a great room for improvement.

\bibliographystyle{plain}
\bibliography{paper}

\newpage
\begin{figure}[h!]
	\begin{minipage}[t]{0.3\textwidth}
		\centering
		\includegraphics[scale=0.28]{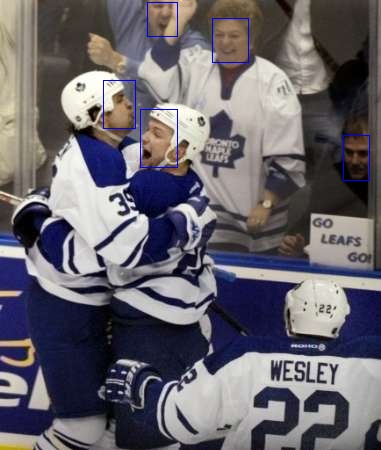}
	\end{minipage}  
	\begin{minipage}[t]{0.3\textwidth}
		\centering  
		\includegraphics[scale=0.34]{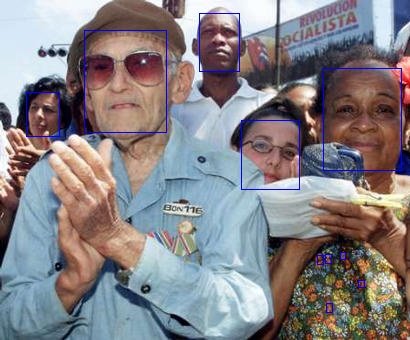} 
	\end{minipage}  
	\begin{minipage}[t]{0.3\textwidth}
		\centering  
		\includegraphics[scale=0.34]{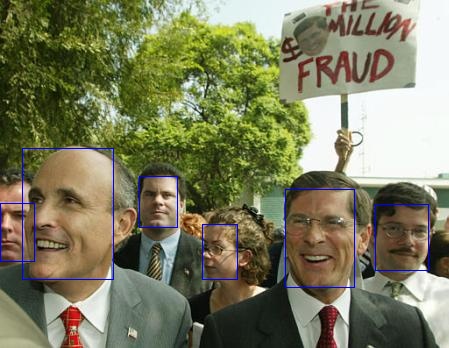}
	\end{minipage}
	\\\\
	\begin{minipage}[t]{0.3\textwidth}
		\centering  
		\includegraphics[scale=0.32]{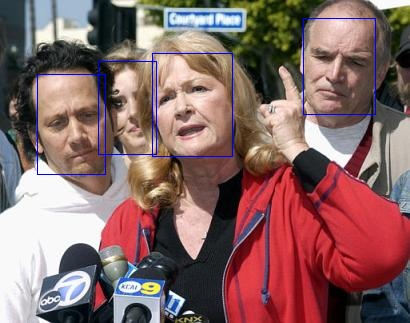}
	\end{minipage}
	\begin{minipage}[t]{0.3\textwidth}
		\centering  
		\includegraphics[scale=0.36]{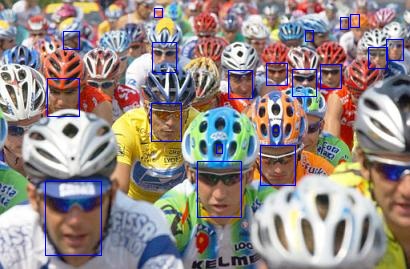}
	\end{minipage}
	\begin{minipage}[t]{0.35\textwidth}
		\centering  
		\includegraphics[scale=0.35]{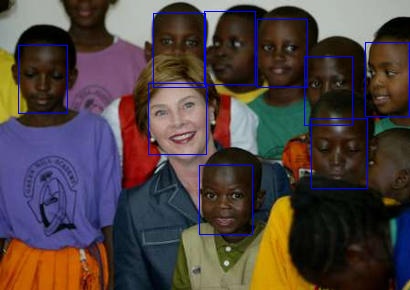}
	\end{minipage}
	\\\\
	\begin{minipage}[t]{0.25\textwidth}
		\centering  
		\includegraphics[scale=0.35]{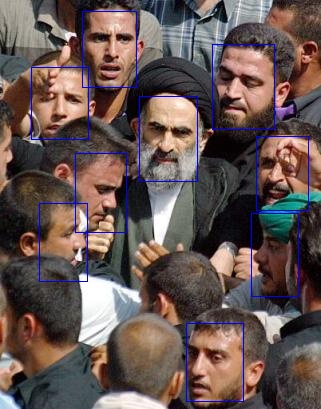}
	\end{minipage}
	\begin{minipage}[t]{0.4\textwidth}
		\centering  
		\includegraphics[scale=0.51]{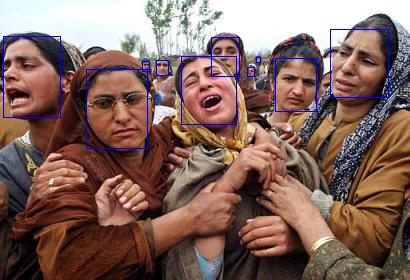}
	\end{minipage}
	\begin{minipage}[t]{0.3\textwidth}
		\centering  
		\includegraphics[scale=0.31]{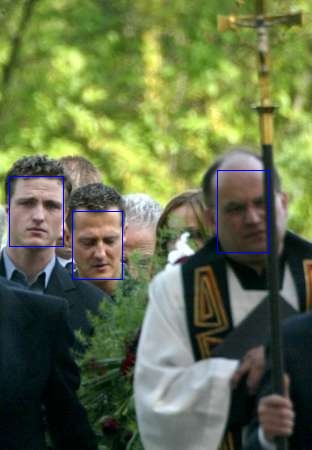}
	\end{minipage}
	\\\\
	\begin{minipage}[t]{0.4\textwidth}
		\centering  
		\includegraphics[scale=0.55]{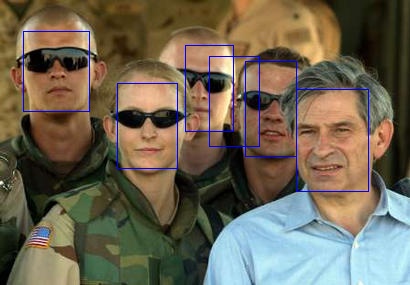}
	\end{minipage}
	\begin{minipage}[t]{0.6\textwidth}
		\centering  
		\includegraphics[scale=0.5]{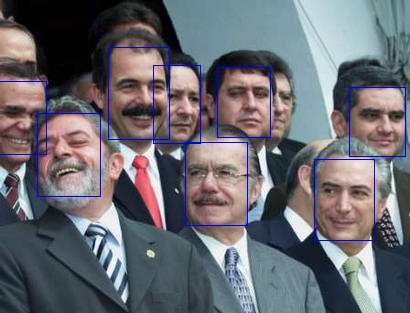}
	\end{minipage}
	\captionsetup{font={scriptsize}}  
	\caption{Some results detected by our FHEDN in FDDB} 
\end{figure}

\begin{figure}[h!]
	\begin{minipage}[t]{0.8\textwidth}
		\centering  
		\includegraphics[scale=0.4]{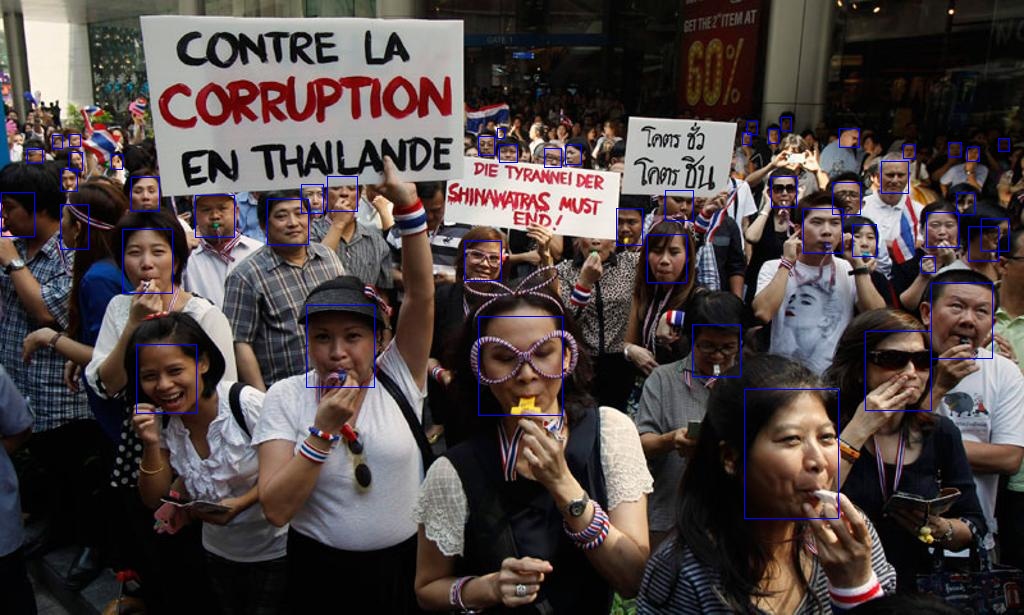}
	\end{minipage}
	\\\\ 
	\begin{minipage}[t]{0.3\textwidth}
		\centering  
		\includegraphics[scale=0.18]{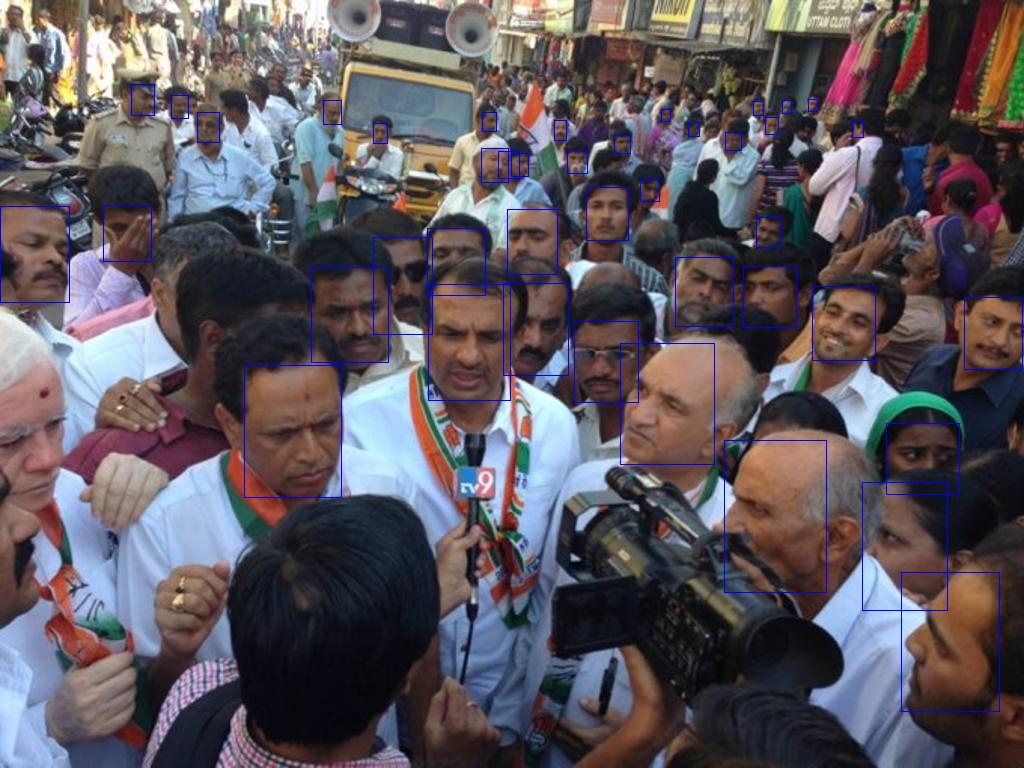}
	\end{minipage}
	\begin{minipage}[t]{0.7\textwidth}
		\centering  
		\includegraphics[scale=0.2]{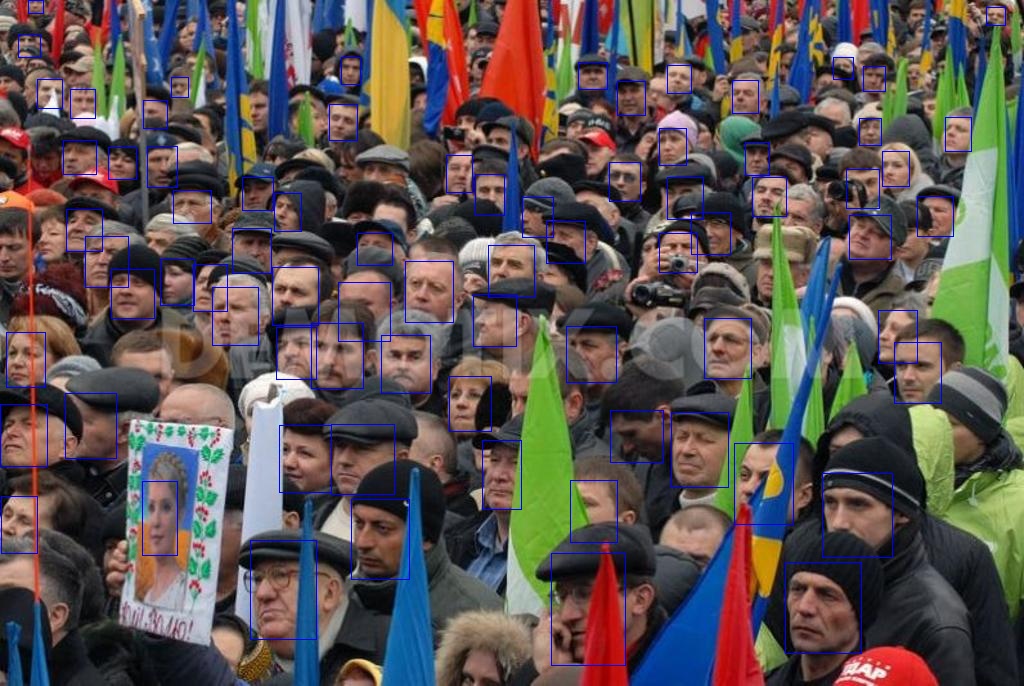} 
	\end{minipage} 
	\\\\
	\begin{minipage}[t]{0.8\textwidth}
		\centering  
		\includegraphics[scale=0.4]{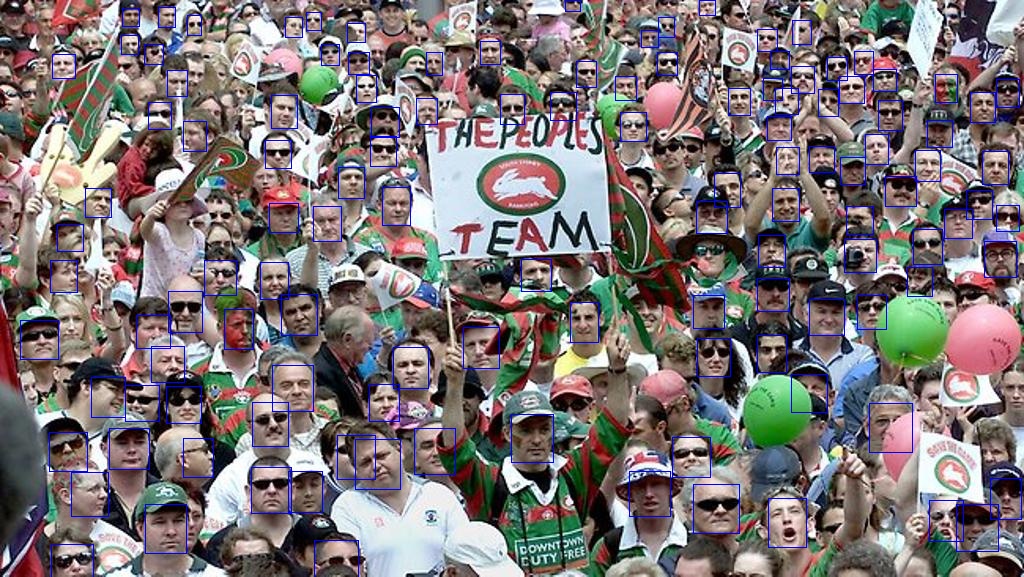}
	\end{minipage}
	
	\captionsetup{font={scriptsize}}  
	\caption{Some results detected by our FHEDN in WIDERFACE} 
\end{figure}

\end{document}